%% file: op_spamACL2011_camera.tex
\documentclass[11pt]{article}
\usepackage{acl-hlt2011}
\usepackage{times}
\usepackage{latexsym}
\usepackage{amsmath}
\usepackage{multirow}
\usepackage{url}
\DeclareMathOperator*{\argmax}{arg\,max}
\usepackage{color}

\newcommand{\inline}[1]{\emph{#1}}

\newcommand{\custitemsep}{\addtolength{\itemsep}{-0.5\baselineskip}}

\title{Finding Deceptive Opinion Spam by Any Stretch of the Imagination}

\author{Myle Ott \hspace*{2em} Yejin Choi \hspace*{2em} Claire Cardie \\
Department of Computer Science \\
Cornell University \\
Ithaca, NY 14853 \\
\url{{myleott,ychoi,cardie}@cs.cornell.edu} \\ \And
Jeffrey T. Hancock \\
Department of Communication \\
Cornell University \\
Ithaca, NY 14853 \\
\url{jth34@cornell.edu} \\}

\begin{document}
\maketitle
\vspace*{-0.5in}
\begin{abstract}
Consumers increasingly rate, review and research products online~\cite{Jansen:10,Litvin:08}. Consequently, websites containing consumer reviews are becoming targets of \emph{opinion spam}. While recent work has focused primarily on manually identifiable instances of opinion spam, in this work we study \emph{deceptive opinion spam}---fictitious opinions that have been deliberately written to sound authentic. Integrating work from psychology and computational linguistics, we develop and compare three approaches to detecting deceptive opinion spam, and ultimately develop a classifier that is nearly 90\% accurate on our \emph{gold-standard} opinion spam dataset. Based on feature analysis of our learned models, we additionally make several theoretical contributions, including revealing a relationship between deceptive opinions and imaginative writing.
\end{abstract}

\section{Introduction}
\label{sec:intro}

With the ever-increasing popularity of review websites that feature user-generated opinions (e.g., TripAdvisor\footnote{\url{http://tripadvisor.com}} and Yelp\footnote{\url{http://yelp.com}}), there comes an increasing potential for monetary gain through \emph{opinion spam}---inappropriate or fraudulent reviews. Opinion spam can range from annoying self-promotion of an unrelated website or blog to deliberate review fraud, as in the recent case\footnote{\url{http://news.cnet.com/8301-1001_3-10145399-92.html}} of a Belkin employee who hired people to write positive reviews for an otherwise poorly reviewed product.\footnote{It is also possible for opinion spam to be negative, potentially in order to sully the reputation of a competitor.}

While other kinds of spam have received considerable computational attention, regrettably there has been little work to date (see Section~\ref{sec:rel}) on opinion spam detection. Furthermore, most previous work in the area has focused on the detection of \textsc{disruptive opinion spam}---uncontroversial instances of spam that are easily identified by a human reader, e.g., advertisements, questions, and other irrelevant or non-opinion text~\cite{Jindal:08}. And while the presence of disruptive opinion spam is certainly a nuisance, the risk it poses to the user is minimal, since the user can always choose to ignore it.

We focus here on a potentially more insidious type of opinion spam: \textsc{deceptive opinion spam}---fictitious opinions that have been deliberately written to sound authentic, in order to deceive the reader. For example, one of the following two hotel reviews is truthful and the other is \emph{deceptive opinion spam}:
\begin{enumerate}\custitemsep \footnotesize
\item I have stayed at many hotels traveling for both business and pleasure and I can honestly stay that The James is tops. The service at the hotel is first class. The rooms are modern and very comfortable. The location is perfect within walking distance to all of the great sights and restaurants. Highly recommend to both business travellers and couples.
\item My husband and I stayed at the James Chicago Hotel for our anniversary. This place is fantastic! We knew as soon as we arrived we made the right choice! The rooms are BEAUTIFUL and the staff very attentive and wonderful!! The area of the hotel is great, since I love to shop I couldn't ask for more!! We will definatly be back to Chicago and we will for sure be back to the James Chicago.
\end{enumerate}

Typically, these deceptive opinions are neither easily ignored nor even identifiable by a human reader;\footnote{The second example review is deceptive opinion spam.} consequently, there are few good sources of labeled data for this research. Indeed, in the absence of gold-standard data, related studies (see Section~\ref{sec:rel}) have been forced to utilize ad hoc procedures for evaluation. In contrast, one contribution of the work presented here is the creation of the first large-scale, publicly available\footnote{Available by request at: \url{http://www.cs.cornell.edu/~myleott/op_spam}} dataset for deceptive opinion spam research, containing 400 truthful and 400 \emph{gold-standard} deceptive reviews.

To obtain a deeper understanding of the nature of deceptive opinion spam, we explore the relative utility of three potentially complementary framings of our problem. Specifically, we view the task as: (a) a standard \emph{text categorization} task, in which we use $n$-gram--based classifiers to label opinions as either deceptive or truthful~\cite{Joachims:98,Sebastiani:02}; (b) an instance of \emph{psycholinguistic deception detection}, in which we expect deceptive statements to exemplify the psychological effects of lying, such as increased negative emotion and psychological distancing~\cite{Hancock:08,Newman:03}; and, (c) a problem of \emph{genre identification}, in which we view deceptive and truthful writing as sub-genres of imaginative and informative writing, respectively~\cite{Biber:99,Rayson:01}.

We compare the performance of each approach on our novel dataset. Particularly, we find that machine learning classifiers trained on features traditionally employed in (a) psychological studies of deception and (b) genre identification are both outperformed at statistically significant levels by $n$-gram--based text categorization techniques. Notably, a combined classifier with both $n$-gram and psychological deception features achieves nearly 90\% cross-validated accuracy on this task. In contrast, we find deceptive opinion spam detection to be well beyond the capabilities of most human judges, who perform roughly at-chance---a finding that is consistent with decades of traditional deception detection research~\cite{Bond:06}.

Additionally, we make several theoretical contributions based on an examination of the feature weights learned by our machine learning classifiers. Specifically, we shed light on an ongoing debate in the deception literature regarding the importance of considering the context and motivation of a deception, rather than simply identifying a universal set of deception cues. We also present findings that are consistent with recent work highlighting the difficulties that liars have encoding spatial information~\cite{Vrij:09}. Lastly, our study of deceptive opinion spam detection as a genre identification problem reveals relationships between deceptive opinions and imaginative writing, and between truthful opinions and informative writing.

The rest of this paper is organized as follows: in Section~\ref{sec:rel}, we summarize related work; in Section~\ref{sec:data}, we explain our methodology for gathering data and evaluate human performance; in Section~\ref{sec:auto}, we describe the features and classifiers employed by our three automated detection approaches; in Section~\ref{sec:resdisc}, we present and discuss experimental results; finally, conclusions and directions for future work are given in Section~\ref{sec:conc}.

\section{Related Work}
\label{sec:rel}

Spam has historically been studied in the contexts of e-mail~\cite{Drucker:02}, and the Web~\cite{Gyongyi:04,Ntoulas:06}. Recently, researchers have began to look at \inline{opinion spam} as well~\cite{Jindal:08,Wu:10,Yoo:09}.

Jindal and Liu~\shortcite{Jindal:08} find that opinion spam is both widespread and different in nature from either e-mail or Web spam. Using product review data, and in the absence of gold-standard deceptive opinions, they train models using features based on the review text, reviewer, and product, to distinguish between \emph{duplicate} opinions\footnote{Duplicate (or near-duplicate) opinions are opinions that appear more than once in the corpus with the same (or similar) text. While these opinions are likely to be deceptive, they are unlikely to be representative of deceptive opinion spam in general. Moreover, they are potentially detectable via off-the-shelf plagiarism detection software.} (considered deceptive spam) and \emph{non-duplicate} opinions (considered truthful). Wu et al.~\shortcite{Wu:10} propose an alternative strategy for detecting deceptive opinion spam in the absence of gold-standard data, based on the distortion of popularity rankings. Both of these heuristic evaluation approaches are unnecessary in our work, since we compare \emph{gold-standard} deceptive and truthful opinions.

Yoo and Gretzel~\shortcite{Yoo:09} gather 40 truthful and 42 deceptive hotel reviews and, using a standard statistical test, manually compare the psychologically relevant linguistic differences between them. In contrast, we create a much larger dataset of 800 opinions that we use to develop and evaluate \emph{automated} deception classifiers.

Research has also been conducted on the related task of \inline{psycholinguistic deception detection}. Newman et al.~\shortcite{Newman:03}, and later Mihalcea and Strapparava~\shortcite{Mihalcea:09}, ask participants to give both their true and untrue views on personal issues (e.g., their stance on the death penalty). Zhou et al.~\shortcite{Zhou:04,Zhou:08} consider computer-mediated deception in role-playing games designed to be played over instant messaging and e-mail. However, while these studies compare $n$-gram--based deception classifiers to a random guess baseline of 50\%, we additionally evaluate and compare two other computational approaches (described in Section~\ref{sec:auto}), as well as the performance of human judges (described in Section~\ref{subsec:human}).

Lastly, automatic approaches to determining \inline{review quality} have been studied---directly~\cite{Weimer:07}, and in the contexts of helpfulness~\cite{Danescu:09,Kim:06,OMahony:09} and credibility~\cite{Weerkamp:08}. Unfortunately, most measures of quality employed in those works are based exclusively on human judgments, which we find in Section~\ref{sec:data} to be poorly calibrated to detecting deceptive opinion spam.

\section{Dataset Construction and Human Performance}
\label{sec:data}

While truthful opinions are ubiquitous online, deceptive opinions are difficult to obtain without resorting to heuristic methods~\cite{Jindal:08,Wu:10}. In this section, we report our efforts to gather (and validate with human judgments) the first publicly available opinion spam dataset with \emph{gold-standard} deceptive opinions.

Following the work of Yoo and Gretzel~\shortcite{Yoo:09}, we compare truthful and deceptive \textbf{positive} reviews for hotels found on TripAdvisor. Specifically, we mine all 5-star truthful reviews from the 20 most popular hotels on TripAdvisor\footnote{TripAdvisor utilizes a proprietary ranking system to assess hotel popularity. We chose the 20 hotels with the greatest number of reviews, irrespective of the TripAdvisor ranking.} in the Chicago area.\footnote{It has been hypothesized that popular offerings are less likely to become targets of deceptive opinion spam, since the relative impact of the spam in such cases is small~\cite{Jindal:08,Lim:10}. By considering only the most popular hotels, we hope to minimize the risk of mining opinion spam and labeling it as truthful.} Deceptive opinions are gathered for those same 20 hotels using Amazon Mechanical Turk\footnote{\url{http://mturk.com}} (AMT). Below, we provide details of the collection methodologies for deceptive (Section~\ref{subsec:amt}) and truthful opinions (Section~\ref{subsec:proc}). Ultimately, we collect 20 truthful and 20 deceptive opinions for each of the 20 chosen hotels (800 opinions total).

\subsection{Deceptive opinions via Mechanical Turk}
\label{subsec:amt}

Crowdsourcing services such as AMT have made large-scale data annotation and collection efforts financially affordable by granting anyone with basic programming skills access to a marketplace of anonymous online workers (known as \emph{Turkers}) willing to complete small tasks.

To solicit gold-standard \textbf{deceptive} opinion spam using AMT, we create a pool of 400 \emph{Human-Intelligence Tasks} (HITs) and allocate them evenly across our 20 chosen hotels. To ensure that opinions are written by unique authors, we allow only a single submission per Turker. We also restrict our task to Turkers who are located in the United States, and who maintain an approval rating of at least 90\%. Turkers are allowed a maximum of 30 minutes to work on the HIT, and are paid one US dollar for an accepted submission.

\input{tables/descstats}

Each HIT presents the Turker with the name and website of a hotel. The HIT instructions ask the Turker to assume that they work for the hotel's marketing department, and to pretend that their boss wants them to write a fake review (as if they were a customer) to be posted on a travel review website; additionally, the review needs to sound realistic and portray the hotel in a positive light. A disclaimer indicates that any submission found to be of insufficient quality (e.g., written for the wrong hotel, unintelligible, unreasonably short,\footnote{\label{foot:short}A submission is considered unreasonably short if it contains fewer than 150 characters.} plagiarized,\footnote{Submissions are individually checked for plagiarism at \url{http://plagiarisma.net}.} etc.) will be rejected.

It took approximately 14 days to collect 400 satisfactory deceptive opinions. Descriptive statistics appear in Table~\ref{table:descstats}. Submissions vary quite dramatically both in length, and time spent on the task. Particularly, nearly 12\% of the submissions were completed in \emph{under one minute}. Surprisingly, an independent two-tailed t-test between the mean length of these submissions ($\bar{\ell}_{t<1}$) and the other submissions ($\bar{\ell}_{t\ge1}$) reveals no significant difference ($p=0.83$). We suspect that these ``\emph{quick}" users may have started working prior to having formally accepted the HIT, presumably to circumvent the imposed time limit. Indeed, the quickest submission took just 5 seconds and contained 114 words.

\input{tables/human}

\subsection{Truthful opinions from TripAdvisor}
\label{subsec:proc}

For truthful opinions, we mine all 6,977 reviews from the 20 most popular Chicago hotels on TripAdvisor. From these we eliminate:
\begin{itemize}\custitemsep
  \item 3,130 non-5-star reviews;
  \item 41 non-English reviews;\footnote{Language is determined using \url{http://tagthe.net}.}
  \item 75 reviews with fewer than 150 characters since, by construction, deceptive opinions are at least 150 characters long (see footnote~\ref{foot:short} in Section~\ref{subsec:amt});
  \item 1,607 reviews written by \emph{first-time authors}---new users who have not previously posted an opinion on TripAdvisor---since these opinions are more likely to contain opinion spam, which would reduce the integrity of our truthful review data~\cite{Wu:10}.
\end{itemize}

Finally, we balance the number of truthful and deceptive opinions by selecting 400 of the remaining 2,124 truthful reviews, such that the document lengths of the selected truthful reviews are similarly distributed to those of the deceptive reviews. Work by Serrano et al.~\shortcite{Serrano:09} suggests that a \emph{log-normal} distribution is appropriate for modeling document lengths. Thus, for each of the 20 chosen hotels, we select 20 truthful reviews from a log-normal (left-truncated at 150 characters) distribution fit to the lengths of the deceptive reviews.\footnote{We use the R package GAMLSS~\cite{Rigby:05} to fit the left-truncated log-normal distribution.} Combined with the 400 deceptive reviews gathered in Section~\ref{subsec:amt} this yields our final dataset of 800 reviews.

\subsection{Human performance}
\label{subsec:human}

Assessing human deception detection performance is important for several reasons. First, there are few other baselines for our classification task; indeed, related studies~\cite{Jindal:08,Mihalcea:09} have only considered a random guess baseline. Second, assessing human performance is necessary to validate the deceptive opinions gathered in Section~\ref{subsec:amt}. If human performance is low, then our deceptive opinions are convincing, and therefore, deserving of further attention.

Our initial approach to assessing human performance on this task was with Mechanical Turk. Unfortunately, we found that some Turkers selected among the choices seemingly at random, presumably to maximize their hourly earnings by obviating the need to read the review. While a similar effect has been observed previously~\cite{Akkaya:10}, there remains no universal solution.

Instead, we solicit the help of three volunteer undergraduate university students to make judgments on a subset of our data. This balanced subset, corresponding to the first fold of our cross-validation experiments described in Section~\ref{sec:resdisc}, contains all 40 reviews from each of four randomly chosen hotels. Unlike the Turkers, our student volunteers are not offered a monetary reward. Consequently, we consider their judgements to be more honest than those obtained via AMT.

Additionally, to test the extent to which the individual human judges are biased, we evaluate the performance of two virtual meta-judges. Specifically, the \textsc{majority} meta-judge predicts ``\emph{deceptive}" when at least two out of three human judges believe the review to be deceptive, and the \textsc{skeptic} meta-judge predicts ``\emph{deceptive}" when \emph{any} human judge believes the review to be deceptive.

Human and meta-judge performance is given in Table~\ref{table:human}. It is clear from the results that human judges are not particularly effective at this task. Indeed, a two-tailed binomial test fails to reject the null hypothesis that \textsc{judge 2} and \textsc{judge 3} perform at-chance ($p=0.003,0.10,0.48$ for the three judges, respectively). Furthermore, all three judges suffer from \emph{truth-bias}~\cite{Vrij:08}, a common finding in deception detection research in which human judges are more likely to classify an opinion as truthful than deceptive. In fact, \textsc{judge 2} classified fewer than 12\% of the opinions as deceptive! Interestingly, this bias is effectively smoothed by the \textsc{skeptic} meta-judge, which produces nearly perfectly class-balanced predictions. A subsequent reevaluation of human performance on this task suggests that the truth-bias can be reduced if judges are given the class-proportions in advance, although such prior knowledge is unrealistic; and ultimately, performance remains similar to that of Table~\ref{table:human}.

Inter-annotator agreement among the three judges, computed using Fleiss' kappa, is 0.11. While there is no precise rule for interpreting kappa scores, Landis and Koch~\shortcite{Landis:77} suggest that scores in the range (0.00, 0.20] correspond to ``\emph{slight agreement}" between annotators. The largest pairwise Cohen's kappa is 0.12, between \textsc{judge 2} and \textsc{judge 3}---a value far below generally accepted pairwise agreement levels. We suspect that agreement among our human judges is so low \emph{precisely because} humans are poor judges of deception~\cite{Vrij:08}, and therefore they perform nearly at-chance respective to one another.

\section{Automated Approaches to Deceptive Opinion Spam Detection}
\label{sec:auto}

We consider three automated approaches to detecting deceptive opinion spam, each of which utilizes classifiers (described in Section~\ref{subsec:class}) trained on the dataset of Section~\ref{sec:data}. The features employed by each strategy are outlined here.

\subsection{Genre identification}
\label{subsec:genre}

Work in computational linguistics has shown that the frequency distribution of \emph{part-of-speech} (POS) tags in a text is often dependent on the genre of the text~\cite{Biber:99,Rayson:01}. In our genre identification approach to deceptive opinion spam detection, we test if such a relationship exists for truthful and deceptive reviews by constructing, for each review, features based on the frequencies of each POS tag.\footnote{We use the Stanford Parser~\cite{Klein:03} to obtain the relative POS frequencies.} These features are also intended to provide a good baseline with which to compare our other automated approaches.

\subsection{Psycholinguistic deception detection}
\label{subsec:ling}

The \emph{Linguistic Inquiry and Word Count} (LIWC) software~\cite{Pennebaker:07} is a popular automated text analysis tool used widely in the social sciences. It has been used to detect personality traits~\cite{Mairesse:07}, to study tutoring dynamics~\cite{Cade:10}, and, most relevantly, to analyze deception~\cite{Hancock:08,Mihalcea:09,Vrij:07}.

While LIWC does not include a text classifier, we can create one with features derived from the LIWC output. In particular, LIWC counts and groups the number of instances of nearly 4,500 keywords into 80 psychologically meaningful dimensions. We construct one feature for each of the 80 LIWC dimensions, which can be summarized broadly under the following four categories:
\begin{enumerate}\custitemsep
  \item Linguistic processes: Functional aspects of text (e.g., the average number of words per sentence, the rate of misspelling, swearing, etc.)
  \item Psychological processes: Includes all social, emotional, cognitive, perceptual and biological processes, as well as anything related to time or space.
  \item Personal concerns: Any references to work, leisure, money, religion, etc. 
  \item Spoken categories: Primarily filler and agreement words.
\end{enumerate}

While other features have been considered in past deception detection work, notably those of Zhou et al.~\shortcite{Zhou:04}, early experiments found LIWC features to perform best. Indeed, the LIWC2007 software used in our experiments subsumes most of the features introduced in other work. Thus, we focus our psycholinguistic approach to deception detection on LIWC-based features.

\subsection{Text categorization}
\label{subsec:text}

In contrast to the other strategies just discussed, our text categorization approach to deception detection allows us to model both content and context with $n$-gram features. Specifically, we consider the following three $n$-gram feature sets, with the corresponding features lowercased and unstemmed: \textsc{unigrams}, \textsc{bigrams$^{+}$}, \textsc{trigrams$^{+}$}, where the superscript $^{+}$ indicates that the feature set subsumes the preceding feature set.

\subsection{Classifiers}
\label{subsec:class}

Features from the three approaches just introduced are used to train Na\"ive Bayes and Support Vector Machine classifiers, both of which have performed well in related work~\cite{Jindal:08,Mihalcea:09,Zhou:08}.

For a document $\vec{x}$, with label $y$, the \inline{Na\"ive Bayes} (NB) classifier gives us the following decision rule:
\begin{equation}\label{eq:NB}
\hat{y} = \argmax_c \Pr( y = c ) \cdot \Pr( \vec{x} \mid y = c )
\end{equation}

When the class prior is \emph{uniform}, for example when the classes are balanced (as in our case), \eqref{eq:NB} can be simplified to the maximum likelihood classifier~\cite{Peng:03}:
\begin{equation}\label{eq:ML}
\hat{y} = \argmax_c \Pr( \vec{x} \mid y = c )
\end{equation}

Under \eqref{eq:ML}, both the NB classifier used by Mihalcea and Strapparava~\shortcite{Mihalcea:09} and the language model classifier used by Zhou et al.~\shortcite{Zhou:08} are equivalent. Thus, following Zhou et al.~\shortcite{Zhou:08}, we use the SRI Language Modeling Toolkit~\cite{Stolcke:02} to estimate individual language models, $\Pr( \vec{x} \mid y = c )$, for truthful and deceptive opinions. We consider all three $n$-gram feature sets, namely \textsc{unigrams}, \textsc{bigrams$^{+}$}, and \textsc{trigrams$^{+}$}, with corresponding language models smoothed using the interpolated Kneser-Ney method~\cite{Chen:96}.

We also train \inline{Support Vector Machine} (SVM) classifiers, which find a high-dimensional separating hyperplane between two groups of data. To simplify feature analysis in Section~\ref{sec:resdisc}, we restrict our evaluation to \emph{linear} SVMs, which learn a weight vector $\vec{w}$ and bias term $b$, such that a document $\vec{x}$ can be classified by:
\begin{equation}\label{eq:SVM}
\hat{y} = sign(\vec{w} \cdot \vec{x} + b)
\end{equation}

We use SVM$^{light}$~\cite{Joachims:99} to train our linear SVM models on all three approaches and feature sets described above, namely \textsc{pos}, \textsc{liwc}, \textsc{unigrams}, \textsc{bigrams$^{+}$}, and \textsc{trigrams$^{+}$}. We also evaluate every combination of these features, but for brevity include only \textsc{liwc+bigrams$^{+}$}, which performs best. Following standard practice, document vectors are normalized to unit-length. For \textsc{liwc+bigrams$^{+}$}, we unit-length normalize \textsc{liwc} and \textsc{bigrams$^{+}$} features individually before combining them.

\section{Results and Discussion}
\label{sec:resdisc}

\input{tables/auto}

The deception detection  strategies described in Section~\ref{sec:auto} are evaluated using a 5-fold \emph{nested} cross-validation (CV) procedure~\cite{Quadrianto:09}, where model parameters are selected for each test fold based on \emph{standard} CV experiments on the training folds. Folds are selected so that each contains \emph{all} reviews from four hotels; thus, learned models are always evaluated on reviews from unseen hotels.

Results appear in Table~\ref{table:auto}. We observe that automated classifiers outperform human judges for every metric, except truthful recall where \textsc{judge 2} performs best.\footnote{As mentioned in Section~\ref{subsec:human}, \textsc{judge 2} classified fewer than 12\% of opinions as deceptive. While achieving 95\% truthful recall, this judge's corresponding precision was not significantly better than chance (two-tailed binomial $p=0.4$).} However, this is expected given that untrained humans often focus on unreliable cues to deception~\cite{Vrij:08}. For example, one study examining deception in online dating found that humans perform at-chance detecting deceptive profiles because they rely on text-based cues that are unrelated to deception, such as second-person pronouns~\cite{Toma:IP}.

\input{tables/POS}

Among the automated classifiers, baseline performance is given by the simple genre identification approach (\textsc{pos$_{\textsc{svm}}$}) proposed in Section~\ref{subsec:genre}. Surprisingly, we find that even this simple automated classifier outperforms most human judges (one-tailed sign test $p=0.06,0.01,0.001$ for the three judges, respectively, on the first fold). This result is best explained by theories of reality monitoring~\cite{Johnson:81}, which suggest that truthful and deceptive opinions might be classified into informative and imaginative genres, respectively. Work by Rayson et al.~\shortcite{Rayson:01} has found strong distributional differences between informative and imaginative writing, namely that the former typically consists of more nouns, adjectives, prepositions, determiners, and coordinating conjunctions, while the latter consists of more verbs,\footnote{\emph{Past participle} verbs were an exception.} adverbs,\footnote{\emph{Superlative} adverbs were an exception.} pronouns, and pre-determiners. Indeed, we find that the weights learned by \textsc{pos$_{\textsc{svm}}$} (found in Table~\ref{table:POS}) are largely in agreement with these findings, notably except for adjective and adverb \emph{superlatives}, the latter of which was found to be an exception by Rayson et al.~\shortcite{Rayson:01}. However, that deceptive opinions contain more superlatives is not unexpected, since deceptive writing (but not necessarily imaginative writing in general) often contains exaggerated language~\cite{Buller:96,Hancock:08}.

Both remaining automated approaches to detecting deceptive opinion spam outperform the simple genre identification baseline just discussed. Specifically, the psycholinguistic approach (\textsc{liwc$_{\textsc{svm}}$}) proposed in Section~\ref{subsec:ling} performs 3.8\% more accurately (one-tailed sign test $p=0.02$), and the standard text categorization approach proposed in Section~\ref{subsec:text} performs between 14.6\% and 16.6\% more accurately. However, best performance overall is achieved by combining features from these two approaches. Particularly, the combined model \textsc{liwc+bigrams$^{+}_{\textsc{svm}}$} is 89.8\% accurate at detecting deceptive opinion spam.\footnote{The result is not significantly better than \textsc{bigrams$^{+}_{\textsc{svm}}$}.}

Surprisingly, models trained only on \textsc{unigrams}---the simplest $n$-gram feature set---outperform all non--text-categorization approaches, and models trained on \textsc{bigrams$^{+}$} perform \emph{even better} (one-tailed sign test $p = 0.07$). This suggests that a universal set of keyword-based deception cues (e.g., \textsc{liwc}) is not the best approach to detecting deception, and a context-sensitive approach (e.g., \textsc{bigrams$^{+}$}) might be necessary to achieve state-of-the-art deception detection performance.

\input{tables/spatial}

To better understand the models learned by these automated approaches, we report in Table~\ref{table:spatial} the top 15 highest weighted features for each class (\emph{truthful} and \emph{deceptive}) as learned by \textsc{liwc+bigrams$^{+}_{\textsc{svm}}$} and \textsc{liwc$_{\textsc{svm}}$}. In agreement with theories of reality monitoring~\cite{Johnson:81}, we observe that truthful opinions tend to include more sensorial and concrete language than deceptive opinions; in particular, truthful opinions are more specific about spatial configurations (e.g., small, bathroom, on, location). This finding is also supported by recent work by Vrij et al.~\shortcite{Vrij:09} suggesting that liars have considerable difficultly encoding spatial information into their lies. Accordingly, we observe an increased focus in deceptive opinions on aspects external to the hotel being reviewed (e.g., husband, business, vacation).

We also acknowledge several findings that, on the surface, are in contrast to previous psycholinguistic studies of deception~\cite{Hancock:08,Newman:03}. For instance, while deception is often associated with negative emotion terms, our deceptive reviews have more positive and fewer negative emotion terms. This pattern makes sense when one considers the goal of our deceivers, namely to create a positive review~\cite{Buller:96}.

Deception has also previously been associated with decreased usage of first person singular, an effect attributed to psychological distancing~\cite{Newman:03}. In contrast, we find increased first person singular to be among the largest indicators of deception, which we speculate is due to our deceivers attempting to enhance the credibility of their reviews by emphasizing their own presence in the review. Additional work is required, but these findings further suggest the importance of moving beyond a universal set of deceptive language features (e.g., \textsc{liwc}) by considering both the contextual (e.g., \textsc{bigrams$^{+}$}) and motivational parameters underlying a deception as well.

\section{Conclusion and Future Work}
\label{sec:conc}

In this work we have developed the first large-scale dataset containing \emph{gold-standard} deceptive opinion spam. With it, we have shown that the detection of deceptive opinion spam is well beyond the capabilities of human judges, most of whom perform roughly at-chance. Accordingly, we have introduced three \emph{automated} approaches to deceptive opinion spam detection, based on insights coming from research in computational linguistics and psychology. We find that while standard $n$-gram--based text categorization is the best individual detection approach, a \emph{combination} approach using psycholinguistically-motivated features and $n$-gram features can perform slightly better.

Finally, we have made several theoretical contributions. Specifically, our findings suggest the importance of considering both the context (e.g., \textsc{bigrams$^{+}$}) and motivations underlying a deception, rather than strictly adhering to a universal set of deception cues (e.g., \textsc{liwc}). We have also presented results based on the feature weights learned by our classifiers that illustrate the difficulties faced by liars in encoding spatial information. Lastly, we have discovered a plausible relationship between deceptive opinion spam and imaginative writing, based on POS distributional similarities.

Possible directions for future work include an extended evaluation of the methods proposed in this work to both negative opinions, as well as opinions coming from other domains. Many additional approaches to detecting deceptive opinion spam are also possible, and a focus on approaches with high deceptive precision might be useful for production environments.

\section*{Acknowledgments}

This work was supported in part by National Science Foundation Grants BCS-0624277, BCS-0904822, HSD-0624267, IIS-0968450, and NSCC-0904822, as well as a gift from Google, and the Jack Kent Cooke Foundation. We also thank, alphabetically, Rachel Boochever, Cristian Danescu-Niculescu-Mizil, Alicia Granstein, Ulrike Gretzel, Danielle Kirshenblat, Lillian Lee, Bin Lu, Jack Newton, Melissa Sackler, Mark Thomas, and Angie Yoo, as well as members of the Cornell NLP seminar group and the ACL reviewers for their insightful comments, suggestions and advice on various aspects of this work.

\bibliography{op_spamACL2011_camera}{}
\bibliographystyle{acl}

\end{document}

%% file: tables/descstats.tex
\begin{table}[t]\footnotesize
\begin{center}
\begin{tabular}{|l|l|}
  \hline
  \multicolumn{2}{|c|}{\textbf{Time spent $t$ (minutes)}} \\
  \hline
  \multirow{3}{*}{All submissions}
    & $count$: 400 \\
    & $t_{min}$: 0.08, $t_{max}$: 29.78 \\
    & $\bar{t}$: 8.06, $s$: 6.32 \\
  \hline
  \hline
  \multicolumn{2}{|c|}{\textbf{Length $\ell$ (words)}} \\
  \hline
  \multirow{2}{*}{All submissions}
    & $\ell_{min}$: 25, $\ell_{max}$: 425 \\
    & $\bar{\ell}$: 115.75, $s$: 61.30 \\
  \hline
  \multirow{3}{*}{Time spent $t$ $< 1$}
    & $count$: 47 \\
    & $\ell_{min}$: 39, $\ell_{max}$: 407 \\
    & $\bar{\ell}$: 113.94, $s$: 66.24 \\
  \hline
  \multirow{3}{*}{Time spent $t$ $\ge 1$}
    & $count$: 353 \\
    & $\ell_{min}$: 25, $\ell_{max}$: 425 \\
    & $\bar{\ell}$: 115.99, $s$: 60.71 \\
  \hline
\end{tabular}
\end{center}
\caption{\label{table:descstats} Descriptive statistics for 400 deceptive opinion spam submissions gathered using AMT.
$s$ corresponds to the sample standard deviation.}
\end{table}

%% file: tables/human.tex
\begin{table*}[t]\footnotesize
\begin{center}
\begin{tabular}{ccc|c|c|c|c|c|c|}
  \cline{4-9}
  & & & \multicolumn{3}{c|}{\textsc{truthful}} & \multicolumn{3}{c|}{\textsc{deceptive}} \\
  \cline{3-9}
  & & \multicolumn{1}{|c|}{\textbf{Accuracy}} & \textbf{P} & \textbf{R} & \textbf{F} & \textbf{P} & \textbf{R} & \textbf{F} \\
  \hline
  \multicolumn{1}{|c|}{\multirow{3}{*}{\textsc{human}}} & \multicolumn{1}{c|}{\textsc{judge 1}} & \textbf{61.9\%} & 57.9 & 87.5 & \textbf{69.7} & 74.4 & 36.3 & 48.7 \\
  \cline{2-9}
  \multicolumn{1}{|c|}{} & \multicolumn{1}{c|}{\textsc{judge 2}} & 56.9\% & 53.9 & \textbf{95.0} & 68.8 & \textbf{78.9} & 18.8 & 30.3 \\
  \cline{2-9}
  \multicolumn{1}{|c|}{} & \multicolumn{1}{c|}{\textsc{judge 3}} & 53.1\% & 52.3 & 70.0 & 59.9 & 54.7 & 36.3 & 43.6 \\
  \hline
  \multicolumn{1}{|c|}{\multirow{2}{*}{\textsc{meta}}} & \multicolumn{1}{c|}{\textsc{majority}} & 58.1\% & 54.8 & 92.5 & 68.8 & 76.0 & 23.8 & 36.2 \\
  \cline{2-9}
  \multicolumn{1}{|c|}{} & \multicolumn{1}{c|}{\textsc{skeptic}} & 60.6\% & \textbf{60.8} & 60.0 & 60.4 & 60.5 & \textbf{61.3} & \textbf{60.9} \\
  \hline
\end{tabular}
\end{center}
\caption{\label{table:human} Performance of three human judges and two meta-judges on a subset of 160 opinions, corresponding to the first fold of our cross-validation experiments in Section~\ref{sec:resdisc}. Boldface indicates the largest value for each column.}
\end{table*}

%% file: tables/auto.tex
\begin{table*}[t]\footnotesize
\begin{center}
\begin{tabular}{ccc|c|c|c|c|c|c|}
  \cline{4-9}
  & & & \multicolumn{3}{c|}{\textsc{truthful}} & \multicolumn{3}{c|}{\textsc{deceptive}} \\
  \hline
  \multicolumn{1}{|c|}{\textbf{Approach}} & \multicolumn{1}{c|}{\textbf{Features}} & \textbf{Accuracy} & \textbf{P} & \textbf{R} & \textbf{F} & \textbf{P} & \textbf{R} & \textbf{F} \\
  \hline
  \multicolumn{1}{|c|}{\textsc{genre identification}} & \multicolumn{1}{c|}{\textsc{pos$_{\textsc{svm}}$}} & 73.0\% & 75.3 & 68.5 & 71.7 & 71.1 & 77.5 & 74.2 \\
  \hline
  \multicolumn{1}{|c|}{\textsc{psycholinguistic}} & \multicolumn{1}{c|}{\multirow{2}{*}{\textsc{liwc$_{\textsc{svm}}$}}} & \multirow{2}{*}{76.8\%} & \multirow{2}{*}{77.2} & \multirow{2}{*}{76.0} & \multirow{2}{*}{76.6} & \multirow{2}{*}{76.4} & \multirow{2}{*}{77.5} & \multirow{2}{*}{76.9} \\
  \multicolumn{1}{|c|}{\textsc{deception detection}} & \multicolumn{1}{c|}{} & & & & & & & \\
  \hline
\multicolumn{1}{|c|}{\multirow{7}{*}{\textsc{text categorization}}} & \multicolumn{1}{c|}{\textsc{unigrams$_{\textsc{svm}}$}} & 88.4\% & 89.9 & 86.5 & 88.2 & 87.0 & 90.3 & 88.6 \\
  \cline{2-9}
  \multicolumn{1}{|c|}{} & \multicolumn{1}{c|}{\textsc{bigrams$^{+}_{\textsc{svm}}$}} & 89.6\% & 90.1 & 89.0 & 89.6 & 89.1 & 90.3 & 89.7 \\
  \cline{2-9}
  \multicolumn{1}{|c|}{} & \multicolumn{1}{c|}{\textsc{liwc+bigrams$^{+}_{\textsc{svm}}$}} & \textbf{89.8}\% & 89.8 & \textbf{89.8} & \textbf{89.8} & \textbf{89.8} & 89.8 & \textbf{89.8} \\
  \cline{2-9}
  \multicolumn{1}{|c|}{} & \multicolumn{1}{c|}{\textsc{trigrams$^{+}_{\textsc{svm}}$}} & 89.0\% & 89.0 & 89.0 & 89.0 & 89.0 & 89.0 & 89.0 \\
  \cline{2-9}
  \multicolumn{1}{|c|}{} & \multicolumn{1}{c|}{\textsc{unigrams}$_{\textsc{nb}}$} & 88.4\% & \textbf{92.5} & 83.5 & 87.8 & 85.0 & \textbf{93.3} & 88.9 \\
  \cline{2-9}
  \multicolumn{1}{|c|}{} & \multicolumn{1}{c|}{\textsc{bigrams$^{+}_{\textsc{nb}}$}} & 88.9\% & 89.8 & 87.8 & 88.7 & 88.0 & 90.0 & 89.0 \\
  \cline{2-9}
  \multicolumn{1}{|c|}{} & \multicolumn{1}{c|}{\textsc{trigrams$^{+}_{\textsc{nb}}$}} & 87.6\% & 87.7 & 87.5 & 87.6 & 87.5 & 87.8 & 87.6 \\
  \hline
  \hline
  \multicolumn{1}{|c|}{\multirow{3}{*}{\textsc{human / meta}}} & \multicolumn{1}{c|}{\textsc{judge 1}} & \textbf{61.9}\% & 57.9 & 87.5 & \textbf{69.7} & 74.4 & 36.3 & 48.7 \\
  \cline{2-9}
  \multicolumn{1}{|c|}{} & \multicolumn{1}{c|}{\textsc{judge 2}} & 56.9\% & 53.9 & \textbf{95.0} & 68.8 & \textbf{78.9} & 18.8 & 30.3 \\
  \cline{2-9}
  \multicolumn{1}{|c|}{} & \multicolumn{1}{c|}{\textsc{skeptic}} & 60.6\% & \textbf{60.8} & 60.0 & 60.4 & 60.5 & \textbf{61.3} & \textbf{60.9} \\
  \hline
\end{tabular}
\end{center}
\caption{\label{table:auto} Automated classifier performance for three approaches based on nested 5-fold cross-validation experiments. Reported precision, recall and F-score are computed using a micro-average, i.e., from the \emph{aggregate} true positive, false positive and false negative rates, as suggested by Forman and Scholz~\shortcite{Forman:09}. Human performance is repeated here for \textsc{judge 1}, \textsc{judge 2} and the \textsc{skeptic} meta-judge, although they cannot be directly compared since the 160-opinion subset on which they are assessed only corresponds to the first cross-validation fold.}
\end{table*}

%% file: tables/POS.tex
\begin{table*}[t]\footnotesize
\begin{center}
\begin{tabular}{|c|l|c|c|l|c|}
  \hline
  \multicolumn{3}{|c|}{\textsc{truthful/informative}}                                     & \multicolumn{3}{c|}{\textsc{deceptive/imaginative}} \\
  \hline
  \textbf{Category} & \multicolumn{1}{c|}{\textbf{Variant}} & \textbf{Weight}             & \textbf{Category} & \multicolumn{1}{c|}{\textbf{Variant}} & \textbf{Weight} \\
  \hline
  \multicolumn{1}{|c|}{\multirow{4}{*}{\textsc{nouns}}} & Singular & 0.008                & \multicolumn{1}{c|}{\multirow{7}{*}{\textsc{verbs}}} & Base & -0.057 \\
  \cline{2-3}                                                                             \cline{5-6}
  \multicolumn{1}{|c|}{} & Plural & 0.002                                                 & \multicolumn{1}{c|}{} & Past tense & \textbf{0.041} \\
  \cline{2-3}                                                                             \cline{5-6}
  \multicolumn{1}{|c|}{} & Proper, singular & \textbf{-0.041}                             & \multicolumn{1}{c|}{} & Present participle & -0.089 \\
  \cline{2-3}                                                                             \cline{5-6}
  \multicolumn{1}{|c|}{} & Proper, plural & 0.091                                         & \multicolumn{1}{c|}{} & Singular, present & -0.031 \\
  \cline{1-3}                                                                             \cline{5-6}
  \multicolumn{1}{|c|}{\multirow{3}{*}{\textsc{adjectives}}} & General & 0.002            & \multicolumn{1}{c|}{} & Third person & \multirow{2}{*}{\textbf{0.026}} \\
  \cline{2-3}
  \multicolumn{1}{|c|}{} & Comparative & 0.058                                            & \multicolumn{1}{c|}{} & singular, present & \multicolumn{1}{c|}{} \\
  \cline{2-3}                                                                             \cline{5-6}
  \multicolumn{1}{|c|}{} & Superlative & \textbf{-0.164}                                  & \multicolumn{1}{c|}{} & Modal & -0.063 \\
  \cline{1-3}                                                                             \cline{4-6}
  \multicolumn{1}{|c|}{\multirow{1}{*}{\textsc{prepositions}}} & General & 0.064          & \multicolumn{1}{c|}{\multirow{2}{*}{\textsc{adverbs}}} & General & \textbf{0.001} \\
  \cline{1-3}                                                                             \cline{5-6}
  \multicolumn{1}{|c|}{\multirow{1}{*}{\textsc{determiners}}} & General & 0.009           & \multicolumn{1}{c|}{} & Comparative & -0.035 \\
  \cline{1-3}                                                                             \cline{4-6}
  \multicolumn{1}{|c|}{\multirow{1}{*}{\textsc{coord.\ conj.}}} & General & 0.094         & \multicolumn{1}{c|}{\multirow{2}{*}{\textsc{pronouns}}} & Personal & -0.098 \\
  \cline{1-3}                                                                             \cline{5-6}
  \multicolumn{1}{|c|}{\multirow{1}{*}{\textsc{verbs}}} & Past participle & 0.053         & \multicolumn{1}{c|}{} & Possessive & -0.303 \\
  \cline{1-3}                                                                             \cline{4-6}
  \multicolumn{1}{|c|}{\multirow{1}{*}{\textsc{adverbs}}} & Superlative & \textbf{-0.094} & \multicolumn{1}{c|}{\multirow{1}{*}{\textsc{pre-determiners}}} & General & \textbf{0.017} \\
  \cline{1-3}                                                                             \cline{4-6}
\end{tabular}
\end{center}
\caption{\label{table:POS} Average feature weights learned by \textsc{pos}$_{\textsc{svm}}$. Based on work by Rayson et al.~\shortcite{Rayson:01}, we expect weights on the left to be positive (predictive of \emph{truthful} opinions), and weights on the right to be negative (predictive of \emph{deceptive} opinions). Boldface entries are at odds with these expectations. We report average feature weights of \emph{unit-normalized} weight vectors, rather than \emph{raw} weights vectors, to account for potential differences in magnitude between the folds.}
\end{table*}

%% file: tables/spatial.tex
\begin{table}[t]\footnotesize
\begin{center}
\begin{tabular}{ll|ll}
  \hline
  \multicolumn{2}{c|}{\textsc{liwc+bigrams$^{+}_{\textsc{svm}}$}} & \multicolumn{2}{c}{\textsc{liwc$_{\textsc{svm}}$}} \\
  \hline
  \multicolumn{1}{c}{\textsc{truthful}} & \multicolumn{1}{c|}{\textsc{deceptive}} & \multicolumn{1}{c}{\textsc{truthful}} & \multicolumn{1}{c}{\textsc{deceptive}} \\
  \hline
  - & chicago & hear & i \\
  ... & my & number & family \\
  on & hotel & allpunct & perspron \\
  location & ,\_and & negemo & see \\
  ) & luxury & dash & pronoun \\
  allpunct$_{\textsc{liwc}}$ & experience & exclusive & leisure \\
  floor & hilton & we & exclampunct \\
  ( & business & sexual & sixletters \\
  the\_hotel & vacation & period & posemo \\
  bathroom & i & otherpunct & comma \\
  small & spa & space & cause \\
  helpful & looking & human & auxverb \\
  \$ & while & past & future \\
  hotel\_. & husband & inhibition & perceptual \\
  other & my\_husband & assent & feel \\
  \hline
\end{tabular}
\end{center}
\caption{\label{table:spatial} Top 15 highest weighted truthful and deceptive features learned by \textsc{liwc+bigrams$^{+}_{\textsc{svm}}$} and \textsc{liwc$_{\textsc{svm}}$}. Ambiguous features are subscripted to indicate the source of the feature. LIWC features correspond to groups of keywords as explained in Section~\ref{subsec:ling}; more details about LIWC and the LIWC categories are available at \url{http://liwc.net}.}
\end{table}